# A stepped sampling method for video detection using LSTM

**Dengshan Li[1,2,3], Rujing Wang[1,3], Chengjun Xie[1,3]**

[1] Institute of Intelligent Machines, Hefei Institute of Physical Science, Chinese Academy Sciences, Hefei 230031, China
[2] Science Island Branch of Graduate School, University of Science and Technology of China, Hefei 230026, China
[3] Intelligent Agriculture Engineering Laboratory of Anhui Province, Hefei 230031, China

Corresponding author: Rujing Wang (rjwang@iim.ac.cn)

**Abstract**
Artificial neural networks that simulate human achieves great successes. From the perspective of simulating human memory method, we propose a stepped sampler based on the "repeated input". We repeatedly inputted data to the LSTM model stepwise in a batch. The stepped sampler is used to strengthen the ability of fusing the temporal information in LSTM. We tested the stepped sampler on the LSTM built-in in PyTorch. Compared with the traditional sampler of PyTorch, such as sequential sampler, batch sampler, the training loss of the proposed stepped sampler converges faster in the training of the model, and the training loss after convergence is more stable. Meanwhile, it can maintain a higher test accuracy. We quantified the algorithm of the stepped sampler. We assume that, the artificial neural networks have human-like characteristics, and human learning method could be used for machine learning.

## 1 Introduction

The artificial intelligence technology is used in numerous fields today. For example, AlphaGo [1] breaks through the limitations of the exhaustive method, which used in traditional chess algorithms, and specifically processes to the most possible direction. Machine learning is one of the approaches to realize artificial intelligence. Machine learning is a process of simulating human learning [2], i.e., learning from a large amount of data to solve various tasks, such as feature extraction and the classification in classification tasks, encoding and decoding in machine translation tasks, etc. The machine learning technology is used in industrial production, financial analysis, agriculture, etc.

Machine learning model is composed of a series of neural network layers. These neural network layers usually consist of convolutional layers, activation functions, pooling layers, etc. According to the principle of artificial neural networks, these layers are usually relatively fixed, although it is not absolute.

The emergence of convolutional neural networks (CNN) [3] has improved the self-learning ability of neural networks. AlexNet [4] using CNN structure won the ImageNet Large Scale Visual Recognition Challenge (ILSVRC) 2012 championship with a large advantage. For image recognition of complex scenes, the recognition accuracy of computer with deep learning algorithms has surpassed human eye [5]. GoogLeNet [6] expands both the depth and the width of the neural network. The depth expansion of the net can extract higher-level semantic features of the image, and the width expansion can extract more dimensional features, making the network extract features richer and more accurate. VGG net [7] may have better universality, and can be used in a variety of different scenes. ResNet [8] uses shortcut to reduce the loss of the input information during the network forward propagation, and the network can detect both low-level and high-level semantic features simultaneously, which improves the detection accuracy. Moreover, ResNet reduces the gradient disappearance when the network is deepened, which seems to be related with the reduction of the information loss. Compared with ResNet, DenseNet [9] adds more shortcut connections, which further reduces the input information loss and gradient disappearance.

The learning effect of the traditional neural network is not good, when processing the data with temporal informations. Recurrent Neural Network (RNN) [10] is used to process the temporal information data. RNN takes the output of the previous time period as the input of the next time period, effectively using the temporal information of the input sequence. RNN can be summarized as:

$$h_t = \sigma(x_t \times w_{xt} + h_{t-1} \times w_{ht} + b) \quad (1)$$

where $\sigma$ is the activation function, it is usually a sigmoid function. $x_t$ and $w_{xt}$ are the input data and input weight of the current time period, and $h_{t-1}$ and $w_{ht}$ are the output and weight of the previous time period. As shown in Figure 1, the $h_{t-1}$ outputted from the previous time period is simultaneously outputted to the next time period.

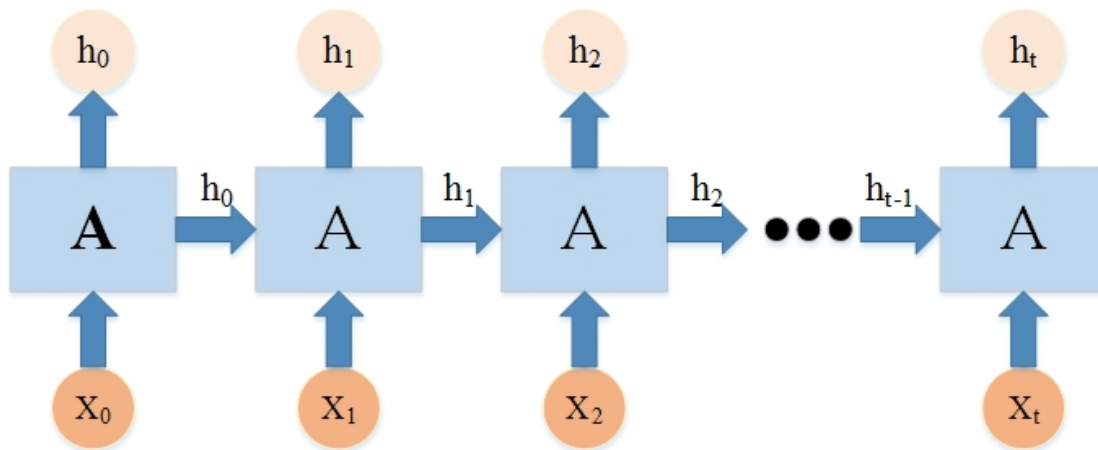

**Fig. 1** The illustration of RNN

RNN sometimes may have the problem of gradient disappearance or gradient explosion. Hochreiter et al. [11] proposed LSTM on the basis of RNN. LSTM adds gates to RNN, thus it can effectively avoid the problem of gradient disappearance or explosion. These gates include the forgetting gates, the input gates, and the output gates. The forgetting gate is the most important among them. The theory of LSTM may simulate the memory process of human brain. Human brain selectively forgets some information for learning better.

The above-mentioned networks develop the detection speed or accuracy from the perspective of the network structure. Another approaches to optimize the deep learning training, without changing the network structure. Mishkin et al. [12] proposed a weight initialization strategy of the training, which can generate a better test accuracy. Ioffe el al. [13] developed Batch Normalization to make the data distribution of the network become normal distribution. The proposed approach can reduce the differences between the data, and accelerate the convergence of the model training. Yu et al. [14] used the frame aggregations between the network layers. The proposed model enhances the deep convolutional neural network (DNN).

Consider that one of the principles of neural networks should be learned from biological neural networks, for those artificial neural networks with the memory effects, such as LSTM, learning from the memory method of human, which is the repeated input and timely review, we study the effect of this method with repeated input on LSTM detection results, without considering changing the LSTM network structure.

In this study, we learn the effect of the proposed input method on neural networks with memory, such as LSTM. Specifically, it is to repeatedly input training data by simulating the "repeated input" memory method of the human brain, in a stepped sampling way.

The structure of the paper is as follows. Section 2 reviews the related work of sampling. Section 3 describes the proposed sampling method in detail, and analyzes the principle of the proposed method.

Section 4 details the experiment, Section 5 and Section 6 are the discussion and conclusion.

## 2 Related Work

The sampler is to preprocess the data through a specific algorithm during training, including batching, or extracting the data via a certain probability, etc. In the literatures, most of the algorithms on the sampler are to extract the data through a probability value. The probability value may be fixed or variable. The variable probability is generated by algorithms.

Gibbs sampling is one of the earlier data sampling algorithms, which is proposed by Geman et al. [15] in 1984. Gibbs sampling is to make the probability of the data sample approximately equal to the required probability distribution via iterations. Gibbs sampling randomly selects data from an initial input sequence, and iterates according to the specified conditional probabilities, which are related to the required probability distribution of the final sampling data. After iterations, Gibbs sampling generates data which is consistent with the required probability distribution.

Hu et al. [16] used neural networks to generate a sampler, which transfer the initial data distribution to the target distribution. The method can generate the sampling data at the same time of training. This method works with the un-normalized probability density function.

Lawson et al. [17] trained neural network models by using truncated rejection sampling, self-normalized importance sampling, and Hamiltonian importance sampling. The models perform better than the previous proposed methods in the experiments of the paper. The models show lower variational bounds of the sampled data.

Wang et al. [18] used Generative Adversarial Nets (GAN) [19] to generate the negative samples. The approach is the first to combine GAN with the negative sampling method, which improves the training effect of the streaming recommend system.

Chen et al. [20] invented an online classifier-sampler communication approach named SampleAhead. The method generates the initial sampling probability of the input data by another sampler, and feeds the sampling probability back to the classifier in real time during training, in order to adjust and improve the prediction of the classifier.

Chu et al. [21] proposed a novel sampler that can sample both the positive and the negative data from the input data sequences, so as to let the classifier utilize the "Regions of Interests" and the "background" of the data. The sampler is used in the few-shot image classifier, which uses the reinforcement learning method. The reinforcement learning algorithm [22] needs to continuously select the regions of interests from the images, subsequently to recognize the content of the Regions of Interests. Sampling these Regions of Interests can improve the efficiency of reinforcement learning, for the reason of the reduction of the input samples. The maximum entropy sampler proposed in the literature [21] can estimate the positive and negative samples, and filter the Regions of Interests of the input image. The experiment in the paper proves the effectiveness of this method.

Liao [23] adopted quasi random numbers on Gibbs sampler to reduce the variance. Usually, the quasi random numbers could not be used in Gibbs sampler, since the sampling is not independent. The algorithm first generates a quasi random numbers from a uniform distribution sequence, and input the generated numbers to the conditional distribution of Gibbs sampler as the condition. A layer of Gibbs sampler is deleted to fit for the change. The approach can enhance Gibbs sampler effectively.

Bouncy Particle Sampler (BPS) [24] is a Markov process, which has a piecewise linear trajectory when sampling. The approach has a significant effect on big data detection. Pakman et al. [25] introduced bias to BPS, the experiment results show that the proposed approach outperforms than the traditional BPS.

Remelli et al. [26] developed an approach which can sample the surface points. The proposed method uses ResNet [8] and 3D convolution [27] to extract the features of point cloud, and uses transposed 3D convolution and pooling layer as the sampler. This method requires less than half of the parameters when sampling the same number of cloud points, compared with the previous approaches.

Ben-Nun et al. [28] used distributed data samplers on distributed deep neural networks (DNN) to establish a high-performance deep learning system. Since a distributed system is used to train the deep learning model, and each computer participates in a part of the deep learning, the training scale of the entire system can be set large. As with the distributed database system, the efficiency would be improved compared with the integrated database system, the training efficiency would be higher than the traditional deep learning system. The distributed data sampler can adopt the "shuffle" operator in the literature [28].

To improve the recommendation system, the implicit feedback is essential to recognize the users' preference. Yu et al. [29] proposed Noisy-label Robust Bayesian Point-wise Optimization (NBPO) to sample the users' preference data, for the purpose of filtering out the users' true negative samples, by learning a model.

Izmailov et al. [30] constructed a subspace with lower dimensions to reduce the number of the parameters of the deep learning model. Elliptical slice sampling (ESS) was proposed and used in the subspace. The experiments show the superiority of the proposed approach.

# 3 Materials and Methods

Contextual information exists in video sequences generally. First, the temporal information comes out between the previous and the next frames, e.g., the position of an object is nearly the same among adjacent frames, the foreground and background between adjacent frames are similar, etc. Second, there is some relationships within a frame, e.g., flowers and grasses always appear together, etc. As a result, deep learning models with contextual informations may be suitable for video detection, such as RNN, LSTM, etc.

Generally, video detection algorithms always decompose videos into frames before detection, i.e., video detection always operates on the frames. Different from image detection, the video detection model emphasizes the association of temporal information between the previous and the next frames. Therefore, the sampling method in this paper is operating on the frames.

This paper is from the perspective of data input, rather than the neural network structure, and study the impact of the "memory" effect on the temporal sequence neural networks (such as LSTM). The "memory" effect simulates the memory process of human brain, namely repeats the input data in a stepped way. The mode we used was the "wheel tactic" [31] when we recited words, by establishing a novel data sampler in the LSTM model training.

## 3.1 Dataset

We try to experiment the proposed sampler on UCF101 [32], which is a human action recognition video dataset, containing 101 action classes, collected from YouTube.

UCF101 has 13320 videos, which could be divided into five categories: human-object interaction, body movement, human-human interaction, musical instrument playing, and sports movement. UCF101 has a large diversity in camera movement, object appearance and posture, object scale, background of the frames, lighting conditions, etc., making the recognition of the dataset more challenging.

The decompressed files contain 13320 folders, corresponding to the 13320 videos. The name of each folder indicates the class of the video. A class includes multiple folders (videos), and folders with the same name and the same class are distinguished by serial numbers.

## 3.2 Video Temporal Information

We consider that, video temporal information is the correlations of the objects between frames, or within a frame. We also consider that, this kind of correlation could be analogized to the correlation of human knowledge, and machine learning may be similar with human learning. From the point, we derive the equations of video temporal information from Bayes' theorem, and mutual information of information theory.

### 3.2.1 The Temporal Information Between the Frames

Bayes' theorem [33], i.e. a conditional probability, can be described as:

$$P(A|B) = \frac{P(A \cap B)}{P(B)} \quad (2)$$

where $P(A|B)$ is the probability of event $A$ occurring, under the condition that event $B$ occurs. $P(A \cap B)$ is the probability of event $A$ and event $B$ occurring at the same time. $P(B)$ is the probability of event $B$ occurring. $P(B) \neq 0$.

We consider that, Bayes' theorem reflects a temporal correlation, that is, event $B$ occurs first, and then event $A$ occurs. The video frame could be analogized to a kind of Venn diagram, and the objects in the frame could be analogized to the events in the Venn diagram. Thus, we apply this event temporal correlation to the temporal information between frames.

We use the area ratio of object $A$ in the video frame $R(A)$ to replace the occurrence probability of the event. The area ratio is defined as:

$$R(A) = \frac{area\ of\ object\ A}{area\ of\ frame} \quad (3)$$

Refer to Bayes' theorem, according to the sequential relations of the temporal information between frames, i.e., the state of the object in the next frame is derived from the state of the object in the previous frame, the temporal information between the frames of the object $A$ is:

$$T_A(nf|pf) = \frac{R(A_{pf} \cap A_{nf})}{R(A_{pf})} \quad (4)$$

where "$nf$" represents the next frame, "$pf$" represents the previous frame, $T_A(nf|pf)$ denotes the temporal information between the frames of the object $A$, $R(A_{pf} \cap A_{nf})$ represents the overlapping area ratio of the previous frame and the next frame of the object $A$. $R(A_{pf})$ represents the area ratio of the object A in the previous frame.

Similarly, the temporal information between the frames of the object $B$ is:

$$T_B(nf|pf) = \frac{R(B_{pf} \cap B_{nf})}{R(B_{pf})} \quad (5)$$

Then, the temporal information between the frames can be described as:

$$T_{bf} = T_A(nf|pf) + T_B(nf|pf) + \cdots$$
$$= \frac{R(A_{pf} \cap A_{nf})}{R(A_{pf})} + \frac{R(B_{pf} \cap B_{nf})}{R(B_{pf})} + \cdots \quad (6)$$

where $T_{bf}$ denotes the temporal information between the frames.

### 3.2.2 The Temporal Information Within a Frame

The temporal information within a frame reflects the correlations of objects in the same image (frame). The correlations within the same image can be represented by mutual information [34] in information theory. The discrete equation of mutual information is

$$I(X;Y) = \sum_{y \in Y} \sum_{x \in X} p(x,y) log\left(\frac{p(x,y)}{p(x)p(y)}\right) \quad (7)$$

where $p(x,y)$ is the joint probability function of $X$ and $Y$, and $p(x)$ and $p(y)$ are the marginal probability functions of $X$ and $Y$.

Referring to the discrete equation of mutual information, we propose the temporal information within a frame as:

when there is an overlapping area between object $A$ and object $B$ in the frame,

$$T(A \cap B) = \sum_{B \in F} \sum_{A \in F} R(A \cap B) log\left(\frac{R(A \cap B)}{R(A)R(B)}\right) \quad (8)$$

where $T(A \cap B)$ represents the intra-frame temporal information when object $A$ and object $B$ have overlapping areas, $F$ represents all objects in the frame, $R(A \cap B)$ represents the overlapping area ratio of object $A$ and object $B$ in the frame, $R(A)$ and $R(B)$ respectively represent the area ratio of the object $A$ and the object $B$ in the frame.

Correlated objects in a frame generally have overlapping areas. For example, blue sky and white clouds, grass and pets, roads and cars, etc. For the correlation of objects in a frame without overlapping areas, since most of the correlated objects in a frame may have overlapping areas, the mathematical representation of the correlation of objects in a frame without overlapping areas can be weakened by a certain form.

We realize this form by taking the logarithm. Then, its mathematical expression is

$$T(A \cup B) = \Sigma_{B \in F} \sum_{A \in F} \log \left( R(A \cup B) \log \left( \frac{R(A \cup B)}{R(A)R(B)} \right) \right) \quad (9)$$

where $T(A \cup B)$ represents the temporal information in a frame with no overlapping area between object $A$ and object $B$, $F$ represents the all objects in the frame, and $R(A \cup B)$ represents the total area ratio of object $A$ and object $B$ in a frame. $R(A)$ represents the area ratio of the object $A$ in the frame, and $R(B)$ represents the area ratio of the object $B$ in the frame.

Since the above equation uses a logarithmic operator,

when $R(A \cup B) \log \left( \frac{R(A \cup B)}{R(A)R(B)} \right) < 1$,

$\log \left( R(A \cup B) \log \left( \frac{R(A \cup B)}{R(A)R(B)} \right) \right) < 0$;

when $R(A \cup B) \log \left( \frac{R(A \cup B)}{R(A)R(B)} \right) > 1$,

$\log \left( R(A \cup B) \log \left( \frac{R(A \cup B)}{R(A)R(B)} \right) \right) > 0$.

This kind of positive and negative value realizes the reduction of the correlation of non-overlapping objects in a frame. Moreover, such negative values are very small. We consider that, the logarithmic operator is suitable for the correlation of non-overlapping objects in a frame.

Then, the temporal information within the frame can be the sum of the overlapping areas and the non-overlapping areas. It can be expressed in mathematical expression as:

$T_{wf} = T(A \cap B) + T(A \cup B) = \Sigma_{B \in F} \Sigma_{A \in F} R(A \cap B) \log \left( \frac{R(A \cap B)}{R(A)R(B)} \right) + \Sigma_{B \in F} \sum_{A \in F} \log \left( R(A \cup B) \log \left( \frac{R(A \cup B)}{R(A)R(B)} \right) \right) \quad (10)$

where $T_{wf}$ denotes the temporal information within a frame.

### 3.2.3 The Equation of Video Temporal Information

The video temporal information is the sum of the one between the frames and the one within a frame. Then, the temporal information of the frame is:

$$T = T_{bf} + T_{wf} = \frac{R(A_{pf} \cap A_{nf})}{R(A_{pf})} + \frac{R(B_{pf} \cap B_{nf})}{R(B_{pf})} + \cdots$$
$$+ \sum_{B \in F} \sum_{A \in F} R(A \cap B) \log \left( \frac{R(A \cap B)}{R(A)R(B)} \right) \bigg|_{A \cap B} +$$
$$\Sigma_{B \in F} \sum_{A \in F} \log \left( R(A \cup B) \log \left( \frac{R(A \cup B)}{R(A)R(B)} \right) \right) \bigg|_{A \cup B} \quad (11)$$

### 3.3 LSTM

Long Short-Term Memory (LSTM) is a special RNN. LSTM solves the problem that the gradient sometimes disappears when RNN learns a long sequence. To some extent, the structure of LSTM is designed for learning long sequences. The main difference between LSTM and other neural networks is the gate in its unit. Generally, there are forget gate, input gate, and output gate in the unit of LSTM.

The gate of LSTM can be described as the following equations:

$$f_t = \sigma(W_f \cdot [h_{t-1}, x_t] + b_f) \quad (12)$$
$$i_t = \sigma(W_i \cdot [h_{t-1}, x_t] + b_i) \quad (13)$$
$$\widetilde{C}_t = \tanh(W_C \cdot [h_{t-1}, x_t] + b_C) \quad (14)$$
$$C_t = f_t * C_{t-1} + i_t * \widetilde{C}_t \quad (15)$$
$$o_t = \sigma(W_o [h_{t-1}, x_t] + b_o) \quad (16)$$
$$h_t = o_t * \tanh(C_t) \quad (17)$$

where $W_f$ is the weight of the forget gate, $h_{t-1}$ is the previous output, $x_t$ is the input, $b_f$ is the bias of the forget gate, the rest mathematic symbol of other equations may be deduced by analogy. Equation 12 reflects the forget gate, Equation 13 denotes the input gate, and Equation 16 indicates the output gate. LSTM should be based on the principle of simulating the memory process of human brain. human brain selectively remembers and absorbs the input information, rather than processing all the information simultaneously.

The LSTM architecture we used in this paper is to start with a CNN backbone and follow by a LSTM model. The CNN backbone has four convolutional layers. The number of convolution kernels (namely the dimension of the convolution layer) of each convolutional layer is 32, 64, 128, 256. The size of each layer convolution kernel is 5×5, 3×3, 3×3, 3×3, the stride size of the convolution kernel is 2, and the padding is 0. Each convolutional layer is followed by a batch normalization (BN) [13] layer and a ReLU layer. The last part of the CNN model is three fully connected (FC) layers, which use dropout function. The dimensions of the three FC layers are 1024, 768 and 512 respectively.

The LSTM used in the paper is the model existed in PyTorch. The input dimension of LSTM is 512, the hidden layer dimension is 512, and the number of hidden layers is 3. The latter is two fully connected (FC) layers followed by dropout function. The dimension of the FC layers is 256. The dropout rate of the CNN backbone and LSTM are both 0.3.

**3.4 Commonly Used Samplers in PyTorch**

Deep learning model training is started with a sampler generally. The reasons of using sampler are the capacity of the RAMs, or improving the speed of model training, etc. Some samplers add the shuffle operation to improve the detection.

Common sampler in PyTorch [35] include random sampler, weighted random sampler, batch sampler, etc. Random sampling is similar to shuffle operation, which could improve the speed of model training, while maintaining an equivalent level of accuracy [36]. Weighted random sampling could be used in some statistical deep learning models, where the random variables can be expressed by the weights. Batch sampling is nearly the most commonly used, the number of the data in a batch (which is named "batch size" in the frameworks) sent to the model determines the occupied size of the GPU or CPU RAMs during training.

**3.5 The Stepped Sampler**

Analogous to human learning, an important point is repetition. And learning data by sampling should be a kind of "repetition", since the data in each batch can be designed to be entered repeatedly.

We suppose that, "repetition" is so important, not only for human beings, but also for computers. To make computers better use the "repetition", analogizing the way how we recite words, we propose a "stepped" repetition input method, which is in the form of the stepped sampler.

The structure of the proposed stepped sampler is illustrated in Figure 2. It is established on the batch sampler. The stepped sampler divides a batch into some sub-batches. There is a duplication between the previous sub-batch and the next sub-batch. Like human memory, this sampler adopts the principle of "adjacent repetition" [37], namely, the back of the previous sub-batch is the same with the front of the next sub-batch. Due to the human-like learning method, we have achieved good results in the experiment compared with the traditional samplers, which will be described in detail in Section 4.

We regard artificial neural networks as biological neural networks. The previous research is to add memory units to deep learning networks, such as RNN, LSTM, etc. In this paper, we try to study the human learning method on the machine system.

The structure of the stepped sampler shows that, the input data of different batches is partly duplicated. The repeated input seems to increase the redundancy, but the experimental results show that, with our experimental environment, this method can accelerate the convergence of LSTM model. There is a stride between the previous sub-batch and the next sub-batch. The stride size $n$ can be set manually. Section 4 describes the comparative experiments on the sampler with different stride size.

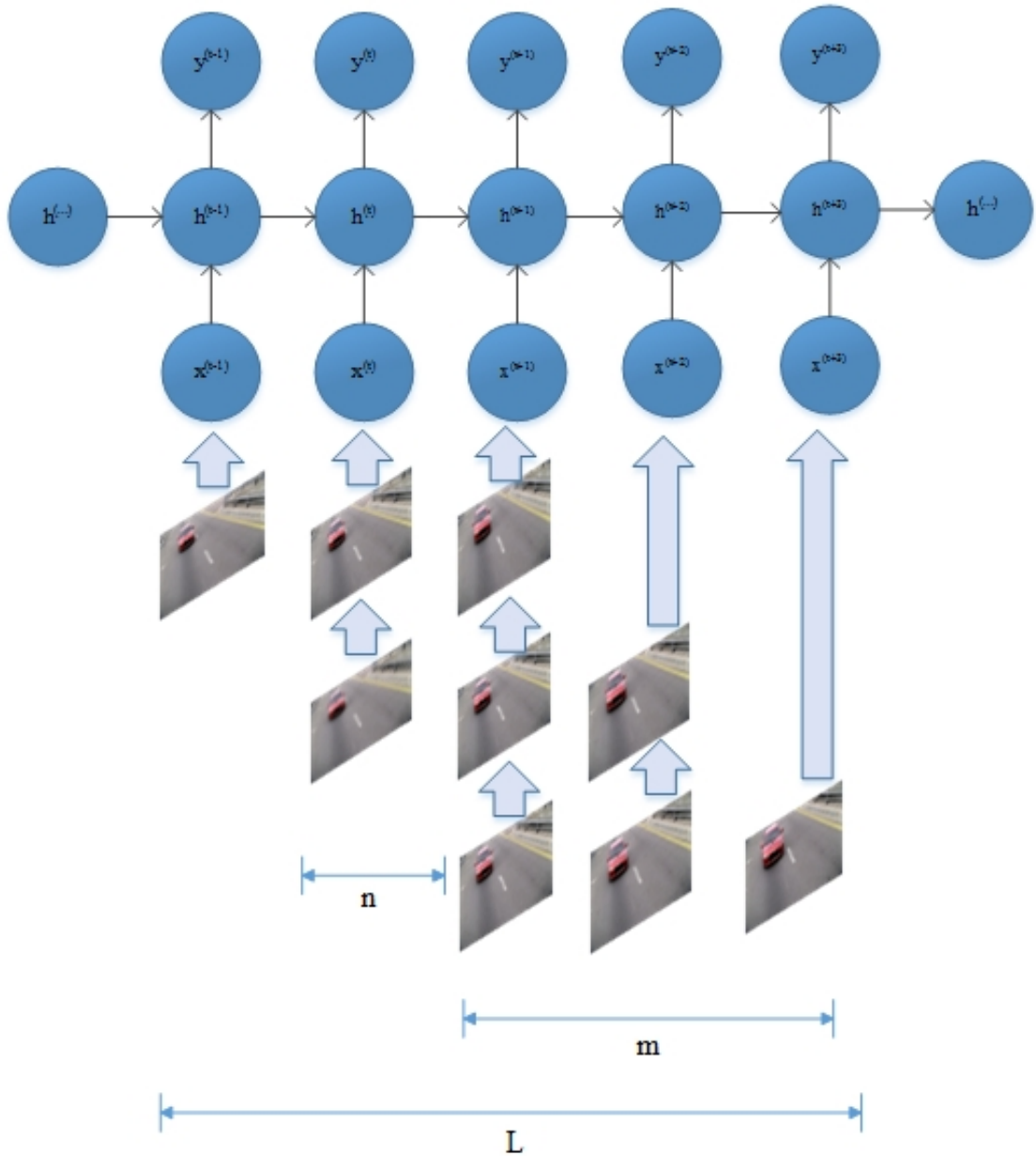

**Fig. 2** The illustration of the proposed stepped sampler. The upper half of the figure is the LSTM network, the lower half describes the workflow of the stepped sampler. The step size *m* is the number of the sub-batch, which is set to 3 in the figure, the step stride *n* is set to 1, the batch size *L* is set to 5 for the illustration. We operated the stepped sampling of the input data within each batch, since we want to strengthen the "repetition".

The stepped sampler can be seen as a kind of data augmentation. In the experiment, we used other data augmentation technologies such as scale transformation, rotation, etc. We consider that the essence of the stepped sampler is a kind of data augmentation.

### 3.6 The Algorithm of the Stepped Sampler

The stepped sampler is designed on the basis of batch sampler. The algorithm is designed to implement stepped sampling within each batch via the batch sampler. The workflow of the algorithm is as follows: the data first goes through the sequential sampler of PyTorch, then, are processed to batches by the batch sampler. Finally, the data in each batch are divided into sub-batches with the same strides by the stepped sampler.

Since batch sampler is operated on the whole data, the proposed stepped sampler is operated on the whole data either. Because the batch sampler is built-in in PyTorch, and the sampler which is built-in in the framework is often faster, we think that using batch sampler speeds up our stepped sampler.

As shown in Figure 2, assuming that the iteration number of the stepped sampler in a batch is $d$, it can be concluded from the figure:

$$L = m + n \times d \quad (18)$$

It can be deduced that, the stepped sampler iteration number per batch, $d$ is:

$$d = \frac{L-m}{n} \quad (19)$$

Especially, if the step stride $n$ was 1 illustrated in Figure 2, it can be seen from Equation 18, the stepped sampler iteration number in a batch, d is:

$$d = L - m \quad (20)$$

Equation 19 and 20 are used as the iteration number within a batch in Algorithm 1. The number of batches is offered by the framework, and the number of epochs is set manually.

---

**Algorithm 1**: The stepped sampler

Input: Dataset, batch size $L$, step size $m$, step stride $n$
Output: Stepped sub-batch of the dataset
1: Initialize the dataset by Sequential sampler of PyTorch.
2: Batch the dataset into batches by Batch sampler of PyTorch.
3: **for** Batch = 1, 2, ⋯, $|len(Batch\ sampler)|$ **do**
4:     Initialize empty set **step_batch[]**.
5:     **for** *idx* = 1, 2, ⋯, L **do**
6:         output the idx-th item *batch[idx]* into **step_batch[]**.
7:         idx += 1.
8:         if *len*(**step_batch[]**) == m
9:             **return step_batch[]**.
10:        Reset **step_batch[]** to empty set.
11:        *idx* = *idx* − *m* + *n*.
12:     **end for**
13: **end for**

---

The algorithm of the proposed sampler is shown in Algorithm 1. The overall idea is to implement the stepped sampler in each batch after the sequential sampler and batch sampler of PyTorch. Line 11 of Algorithm 1 is that, after each previous stepped sub-batch's output, the starting coordinate is moved by n (step stride) data from the starting position of the previous sub-batch.

## 4 Results

### 4.1 Experiment Setup

The system used in the experiment was a workstation with 32 GB CPU RAM and a NVIDIA GeForce 1080ti GPU. The processor was Intel i7 8700, the operating system was Ubuntu 16.04 64 bits. PyTorch version used in the experiment was 1.0.1, Python version was 3.6, Numpy version was 1.20.4, Sklearn version was 0.20.4, and Matplotlib, Pandas, tqdm were implemented as the software environment.

The videos of UCF101 [32] were processed to RGB images frame by frame before the experiment. The frames of a video were put in a folder, and the name of the folder corresponds to the content of the frames. The label of UCF101 is the name of each video.

## 4.2 Training

The detection accuracy evaluation and cross-entropy loss were used for the training of the models. The accuracy evaluation in the experiment used the accuracy score tool in the Sklearn package of Python. The cross-entropy loss used the function of PyTorch. The accuracy and loss were graphically depicted in Figure 4 and 5. The accuracy and loss were computed every epoch. The dataset of UCF101 was split into training set and test set by a ratio of 3:1. After training, an overall accuracy and loss were computed by the test set, to evaluate the performance of the models. The epoch was set to about 150.

For the parameter optimization, we used Adam as the optimization algorithm. Adam is a learning rate-adaptive algorithm, which has the features of less resources occupation and high efficiency, etc. We experimented different batch sizes and step sizes. Which is shown in Figure 2 that, change the size of $L$, $m$ and the step stride $n$ for the experiment.

Our experiment is trained from scratch. The learning rate was set to 0.0001. The momentum was set to 0.01. The operator of batch normalization (BN) and ReLU activation were used after each convolutional layer in the CNN backbone. The CNN backbone is not shown in Figure 2. The data transformation was applied to enhance the network. The input frames were transformed into 256×342 pixels.

## 4.3 Experiment Results

The experiment is to study the detection effect of the proposed sampler, which simulates the human brain memory law, repeating the input. In this experiment, we constructed the stepped sampler as what was illustrated in Section 2.4. We tried to evaluate the proposed sampler from different batch sizes, different step sizes and different step strides. Figure 3 are the visualized results.

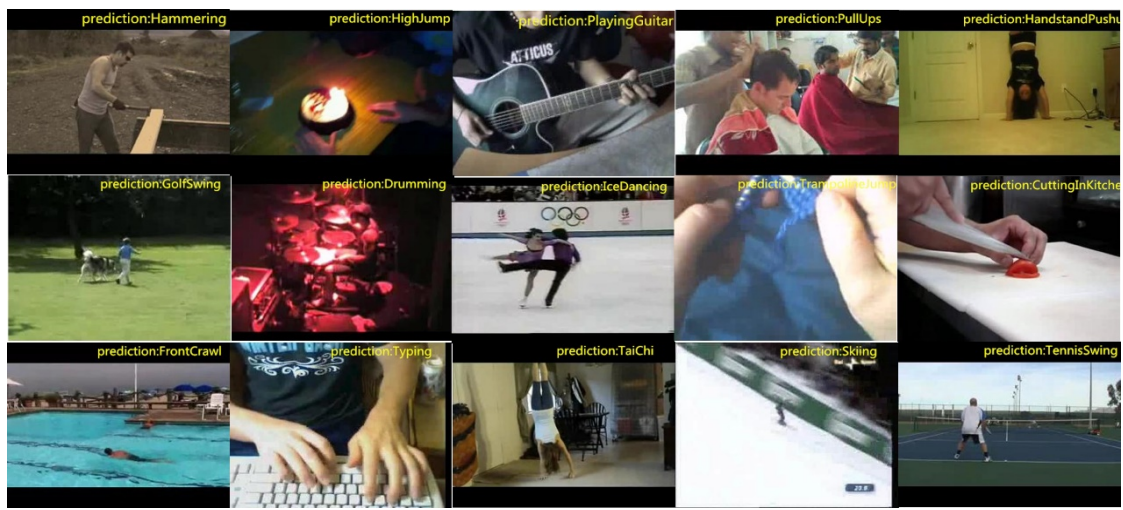

**Fig. 3** The video detection results of UCF101, which uses the proposed stepped sampler. Some frames of the videos are detected wrong, such as the frame of Line 1, Column 2, the groudtruth is "BlowingCandles".

The LSTM we used is existed in PyTorch. We tested the sampler of batch size 25. Figure 4 presents the experimental results. Each subfigure shows the training loss and test accuracy of the models. The difference of the models is only the sampler, for comparing the results only caused by the sampler. Figure 4(a) is the model of traditional sampler, which is the sequential sampler and the batch sampler in PyTorch, and the other in Figure 4 are the models of the proposed stepped sampler. Figure 4 (b), (c), (d), (e), (f) are only different from the step stride for comparing. From Figure 4, we consider that step stride 2 (batch size 25, step size 20, in Figure 4(c)) is the optimal. The training loss in Figure 4(a) has many jitters, even when the epoch is more than 110, while the training loss in the other subfigures are much smoother, and can converge earlier than the traditional batch sampler model (Figure 4(a)). Nonetheless, the test accuracy score of the traditional batch sampler model (Figure 4(a)) is slightly higher. The test accuracy of Figure 4(a) can be 0.656, while the test accuracy of the model with stride 2 stepped sampler (Figure 4(c)) can be 0.603. The test accuracy of the models in Figure 4 is shown in Table 1.

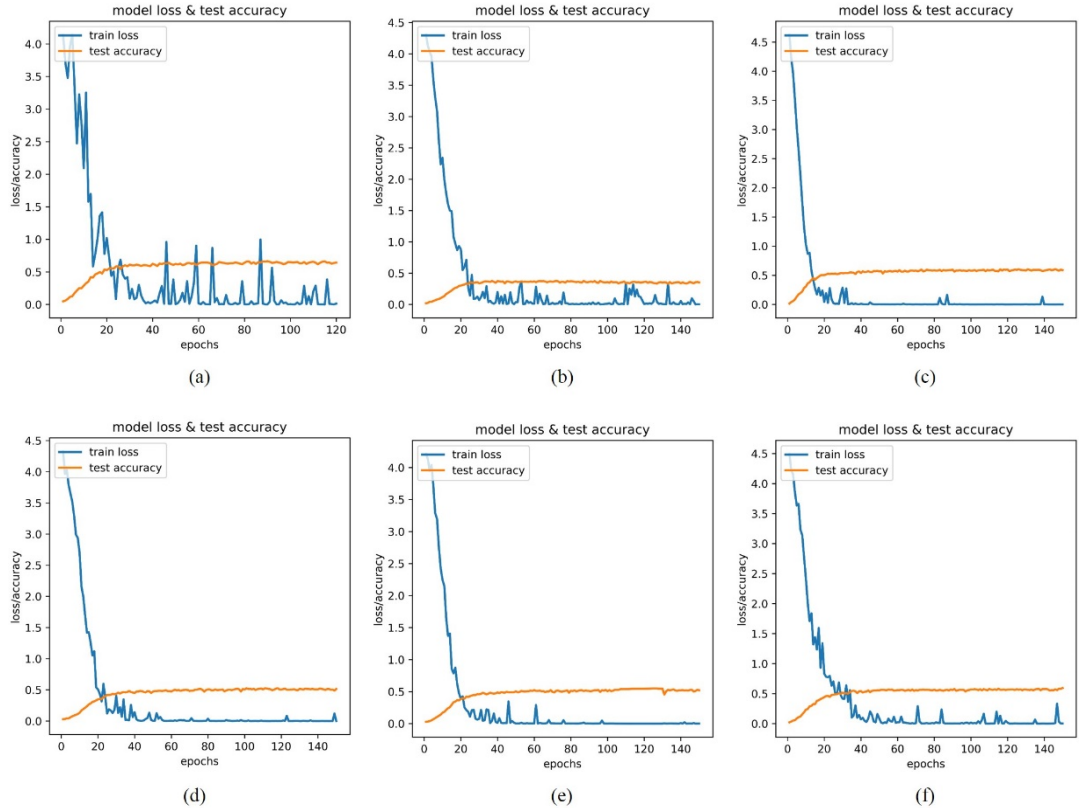

**Fig. 4** The training loss and test accuracy of the traditional sampler LSTM model and the stepped sampler LSTM model. The blue curves denote the training loss, and the yellow curves denote the test accuracy. The above six models are all with batch size 25. Subfigure (a) is a traditional LSTM model, which uses the batch sampler in PyTorch. Subfigure (b) is with the proposed stepped sampler, whose batch size is 25, step size is 20, step stride is 1. Subfigure (c), (d), (e), (f) are with the stepped sampler, whose batch size is 25, step size is 20, step stride is 2, 3, 4, 5, respectively. We want to study the better step stride when the batch size and step size are fixed.

**Table 1** The test accuracy of the six models in Figure 4. "BatchSampler" denotes Figure 4(a), "stride 1 stepped" denotes Figure 4(b), "stride 2 stepped" denotes Figure 4(c), etc.

| model | epoch 10 | epoch 50 | epoch 100 | epoch 120 | epoch 150 |
| --- | --- | --- | --- | --- | --- |
| BatchSampler | 0.289 | 0.621 | 0.650 | 0.642 | 0.635 |
| stride 1 stepped | 0.117 | 0.369 | 0.358 | 0.360 | 0.352 |
| stride 2 stepped | 0.307 | 0.566 | 0.577 | 0.602 | 0.587 |
| stride 3 stepped | 0.133 | 0.471 | 0.506 | 0.505 | 0.514 |
| stride 4 stepped | 0.170 | 0.502 | 0.506 | 0.546 | 0.521 |
| stride 5 stepped | 0.232 | 0.553 | 0.570 | 0.568 | 0.593 |

From Figure 4, the following could be concluded:

**a)** In the model training, LSTM with the stepped sampler converges faster than LSTM with the traditional sampler, and the convergence effect is better, i.e., there is no large jitter after the drop.

**b)** When the batch size and the step size are fixed, the smaller the step stride was, the worse the detection effect became. Similarly, the larger the step stride was, the worse the detection effect became either. If the batch size and the step size are fixed, the detection effect seems to be a normal distribution of the step stride.

**c)** When the epoch is about 70, the training loss and test accuracy tend to converge and stabilize. When the epoch is about 120, the training loss reaches a minimum, and the test accuracy reaches a maximum. More iterations could not lead to better detection results.

**d)** However, LSTM with the traditional sampler whose batch size is 25 has a higher test accuracy on the test set. Although this value is not much higher than the optimal stepped sampler model (Figure 4(c)). The following experiments show that, for LSTM with the traditional sampler, the test accuracy of the model with a larger batch size is higher than the model with a smaller batch size.

**e)** The test accuracy of the traditional LSTM will even decrease after the epoch is 120 (Figure 4(a)), however, the test accuracy of the stepped sampler LSTM is always stable.

From Table 1 we may conclude that:

**a)** for the same batch size, the test accuracy of the LSTM with stepped sampler rises faster than the traditional sampler LSTM. This can also be seen in Figure 4.

**b)** Both the stepped sampler LSTM and the traditional LSTM may approach a relative maximum when the epoch is about 50. The maximum value of the former is 0.566, and the maximum value of the latter is 0.621.

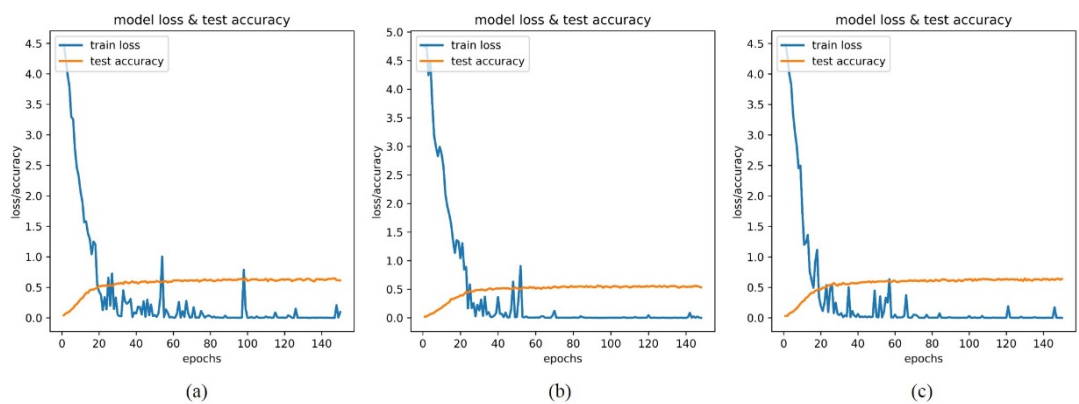

**Fig. 5** The training loss and the test accuracy of the traditional model and our stepped sampler model. The batch size of the three models are all 15. The blue curve denotes the training loss, and the yellow curve denotes the test accuracy. (a) is LSTM with the traditional sampler, batch sampler in PyTorch. (b) and (c) are LSTM with the proposed stepped sampler, with the stepped size 10, when the step stride of (b) is set to 3, and the step stride of (c) is set to 5. The training loss of (c) still converges faster than (a).

Figure 5 is the illustration of the traditional LSTM and the stepped sampler LSTM, when the batch sizes are all set to 15. We can see that, there is a large jitter when the epoch is about 100 in Figure 5(a). Figure 5(b) and Figure 5(c) have no large jitter after the epoch is about 60. The training loss of Figure 5(c) drops faster than Figure 5(a). The test accuracy score of Figure 5(c) is higher than Figure 5(a), which denotes that, the stepped sampler LSTM may have higher test accuracy than the traditional sampler LSTM, when the batch size is 15, step size is 10, step stride is 5 of the stepped sampler LSTM, and the traditional sampler LSTM is with batch size 15. The test accuracy of the three models is shown in Table 2. From Figure 5 we can see that, the training loss of the stepped sampler model still converges faster than the traditional sampler model.

Table 2 The test accuracy of the three models in Figure 5. "BatchSampler" represents Figure 5(a), "stride 3 stepped" represents Figure 5(b), "stride 5 stepped" represents Figure 5(c).

| model | epoch 10 | epoch 50 | epoch 100 | epoch 120 | epoch 150 |
|---|---|---|---|---|---|
| BatchSampler | 0.303 | 0.585 | 0.607 | 0.622 | 0.616 |
| stride 3 stepped | 0.178 | 0.517 | 0.532 | 0.544 | 0.535 |
| **stride 5 stepped** | **0.282** | **0.593** | **0.631** | **0.639** | **0.637** |

In our experiments, most LSTMs with the stepped sampler have a more stable convergence of training loss, compared with the traditional LSTMs based on the same batch size. The stepped sampler LSTM may have a higher test accuracy, which can reach the value of 0.639.

Our test uses the shuffle operation, to make the test results more objective. Since the test data is shuffled, there should be less temporal information between the data, the test results may depend on the sampler and model only, which could reflect the detection effect well.

## 5 Discussion

The previous studies place emphasis on the network structure, however, this paper concentrates on the way of the data input. The purpose is to simulate the repeating learning method of human brain on the temporal information sequences.

The experimental results showed that our method was more effective than the traditional samplers, such as batch sampler, random sampler, etc., on enhancing the learning effect of LSTM. We analyze the reasons as the follows.

As the data are partly repeated inputted, it is equivalent to the timely knowledge review of human brain, which strengthens the memory of the LSTM network, and reduces the information forgetting. LSTM is specially used to process temporal information sequences, therefore, LSTM can selectively memorize the temporal information. Then, using a human-like method to analyze LSTM, we think it is appropriate.

From Figure 4 we can see that, the repeating times of the stepped sampler is not the more the better, as shown in Figure 4(b), the stride is 1, and the convergence speed of the model is not improved much. In addition, the repeating times of the stepped sampler is not the less the better. As shown in Figure 4(f), the convergence speed of the model is even slower. The phenomenon is the same with human learning. Too much repetitive input and too little repetitive input would not improve the learning effect of human. The experiments seem to verify the similarity between machine learning and human learning. Thus, we assume that artificial neural networks have human-like characteristics. We also assume that, one of the ways to realize strong artificial intelligence is still artificial neural networks.

Temporal information also seems to have human-like characteristics. Temporal information is the correlation of temporal sequences, analogous to human learning, it is the correlation of knowledge. In human learning, one of the important learning methods is to use the correlation of knowledge. Then, using the temporal information should also be one of the important learning methods of machine learning.

To our knowledge, imitating the way of preventing forgetting, which is studied in psychology, and combining the computer with psychology, to learn the temporal information sequences, we should be one of the earlier researchers.

The video detection is almost based on the frames. In the literatures, some algorithms are frame-by-frame detection [38], some algorithms extract the key frames to detect [39], and some algorithms use the temporal information between adjacent frames to enhance the detection effect [40], etc. Therefore, the sampling of the data input in this paper is also operated on the frames.

The experimental dataset is UCF101, which is a video dataset of human actions. The dataset has much temporal information between adjacent frames, since there is a strong temporal and spatial correlation between the preceding and following frames. One of the meaning of this correlation is that the scenes of the adjacent frames are always similar.

The stepped sampler is invertible, i.e., the original data sequence can be restored from the sequence after the stepped sampling. The solution is to subtract duplicates from the sampling data.

Whether there were better ways to promote memory effect from psychology, for being used to enhance the LSTM network, or is there any other ways to enhance this "memory effect", those are our next research directions.

# 6 Conclusion

We refer to the human memory rules, and propose the stepped sampler, a repeating input method which uses the timely review approach. In our experiments, this method has a better promotion on the detection effect of LSTM. The experimental results show that, compared with the traditional sampler, the training loss of the stepped sampler converges faster, and is more stable after the convergence, i.e., there is no large jitter after convergence. The test accuracy of the model with the stepped sampler also reaches a high point faster and is more stable either. We analyzed the algorithm of stepped sampler and got several equations.

Another innovative point in the paper is that, we combine computer science with psychology, and use psychological methods to study artificial neural networks. The results show that, there may be a close relationship between biological neural network and artificial neural network, whatever in structure and even in principle. Artificial neural networks seem to have human-like characteristics, and human learning and machine learning seem to have similarities. We assume that, one of the ways to realize strong artificial intelligence is still artificial neural networks.